\documentclass[lettersize,journal]{IEEEtran}
\usepackage{amsmath,amsfonts}
\usepackage{algorithmic}
\usepackage{algorithm}
\usepackage{array}
\usepackage[caption=false,font=normalsize,labelfont=sf,textfont=sf]{subfig}
\usepackage{textcomp}
\usepackage{stfloats}
\usepackage{url}
\usepackage{verbatim}
\usepackage{graphicx}
\usepackage{orcidlink}
\usepackage{multirow}
\usepackage{cite}
\usepackage{bbm}
\usepackage{amssymb}
\usepackage{booktabs}   
\usepackage{tabularx}   
\begin{document}

\title{AstroMind: A High-Fidelity Benchmark for Spacecraft Behavior Reasoning Based on Large Language Models}

\author{Hao Liu$^{\orcidlink{0009-0004-7614-9597}}$,
Siyuan Yang$^{\orcidlink{0000-0003-4681-0431}}$,
Qinglei Hu$^{\orcidlink{0000-0002-5563-310X}}$,
and Dongyu Li$^{\orcidlink{0000-0001-8338-0536}}$
\thanks{Hao Liu is with the Hangzhou International Innovation Institute, Beihang University, Hangzhou 311115, China. E-mail: \url{lh@computer.org}}
\thanks{Siyuan Yang is with the KTH Royal Institute of Technology, Stockholm, Sweden. E-mail: \url{siyuany@kth.se}}
\thanks{Qinglei Hu is with the School of Automation Science and Electrical Engineering, Beihang University, Beijing 100191, China. E-mail: \url{huql_buaa@buaa.edu.cn}}
\thanks{Dongyu Li is with the School of Cyber Science and Technology, Beihang University, Beijing 100191, China. E-mail: \url{dongyuli@buaa.edu.cn}. Corresponding author: Dongyu Li.}}

\maketitle

\begin{abstract}
Understanding why a spacecraft maneuvers---rather than simply that it did---is an increasingly important problem for space domain awareness as Earth orbits grow crowded and contested. Current analysis pipelines are built for detection: they are good at picking up that something happened, less good at reasoning about what it means. AstroMind is a physics-grounded benchmark designed to close that gap. It draws on high-fidelity astrodynamics simulations and real observational constraints, converting them into verifiable reasoning problems across three task types: intent inference, maneuver parameter estimation, and threat assessment. Each scenario includes realistic sensing noise and multi-source textual intelligence at varying reliability levels. Evaluation metrics capture both semantic correctness and quantitative consistency under physical constraints. Benchmarking a suite of open-weight models shows no single model dominates every axis: Qwen3 (32B) leads on intent inference accuracy; QwQ (32B) leads on threat assessment and achieves the lowest median relative error on parsed items; GPT-OSS (20B) produces the strongest judged reasoning quality and extracts the most scalar values for parameter estimation (136 of 241 parsed items). Training data composition and reasoning style matter as much as model size. Structured reasoning prompts help consistently across tested 8B models, with larger gains for those that can already track physical constraints. AstroMind gives the field a shared test for a problem where getting the physics right and reading the tactical situation correctly are both required---neither is sufficient on its own.
\end{abstract}


\section{Introduction}

\begin{figure}[t]
    \centering
    \includegraphics[width=0.98\columnwidth]{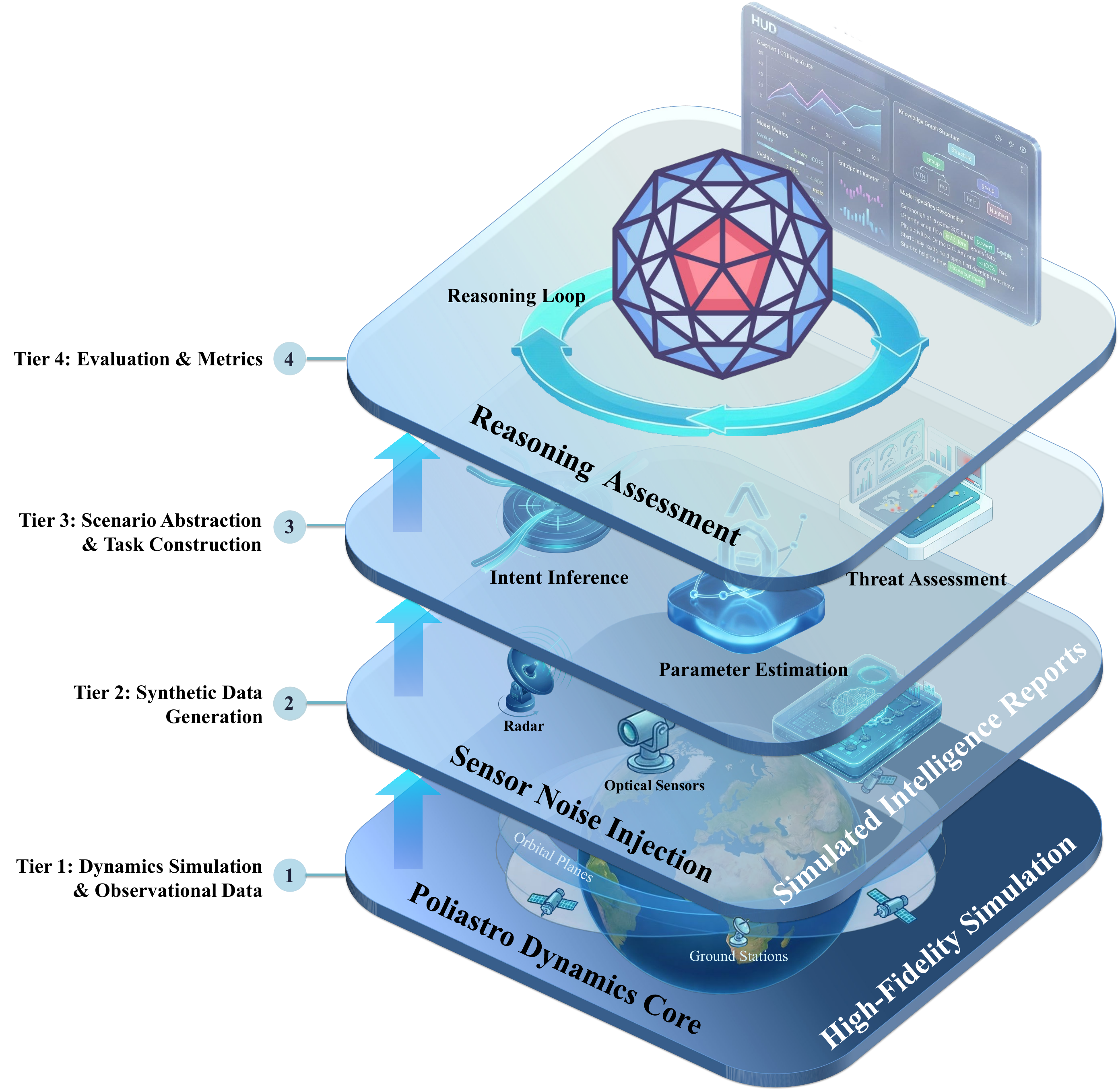}
    \caption{\textbf{The AstroMind Framework.} Four-tier pipeline from astrodynamics to scored LLM tasks. Tier 1 produces physical ground truth via the Poliastro dynamics core; Tier 2 adds sensor noise and simulated intelligence reports; Tier 3 formats the result into structured tasks (Intent Inference, Parameter Estimation, Threat Assessment); Tier 4 scores responses with multi-dimensional metrics and the Reasoning Loop.}
    \label{fig:framework}
\end{figure}

\IEEEPARstart{E}{arth} orbit is getting crowded. The proliferation of large satellite constellations has fundamentally altered the near-Earth environment \cite{esa2025space, McDowell2020TheLE}: collision risk is rising, space traffic management is growing harder, and a dwindling fraction of tracked objects are actually functional satellites. The Kessler Syndrome is no longer just a theoretical concern; cascading collisions could render critical orbital bands unusable for generations \cite{kessler1978collision}. Against that backdrop, analysts face a more immediate problem: distinguishing a routine orbit correction from a maneuver that carries strategic significance.

That distinction has pushed the field from Space Situational Awareness (SSA), which mainly asks \textit{where} objects are, toward Space Domain Awareness (SDA), which asks \textit{why} they moved. The shift matters most in what analysts have taken to calling the ``Congested, Contested, and Competitive'' space environment, where ambiguous proximity operations can carry both operational and security implications. Non-cooperative Rendezvous and Proximity Operations (RPO) sit at the center of this interpretive challenge.

Recent events put a sharp edge on the problem. Russia's Luch/Olymp-K satellites have performed close-approach maneuvers near commercial communications satellites; China's Shijian series has demonstrated RPO capabilities that blur the boundary between on-orbit servicing and something more adversarial \cite{SWF_Counterspace2025, CSIS_SpaceThreat2025}. Making sense of such events requires more than trajectory analysis --- it requires integrating technical context, mission history, and geopolitical background, a reasoning task far beyond the reach of any threshold-crossing algorithm.

The dominant analytical tools today are still rooted in signal processing. Kalman-filter-based residual analysis can reliably detect that a maneuver occurred, but it cannot say why \cite{RAND_AI_SDA_2024}. Supervised machine learning improved detection accuracy and cut false-positive rates, yet it left the fundamental limitation intact: these methods classify statistical patterns, not intentions \cite{Goulet_Intent_2025, Tsaprailis_Survey_2024}. They answer \textit{what happened}, not \textit{what it means}.

Large language models (LLMs) offer a different kind of tool. Because they can combine structured numerical data with unstructured text, including news coverage, operator statements, and intelligence summaries, they are plausibly positioned to bridge orbital mechanics and strategic interpretation \cite{Jin_TimeLLM_2024}. Whether they actually can make that leap, and under what conditions, is an empirical question. Answering it requires a benchmark purpose-built for the task.

No such benchmark currently exists. The LLM evaluation landscape is rich in general-knowledge tests \cite{Hendrycks_MMLU_2021} and growing in specialized domains like medicine \cite{Jin_MedQA_2021} and law \cite{Guha_LegalBench_2023}, but astrodynamics has been left out entirely. Without a physics-grounded, reproducible evaluation framework, there is no principled way to know whether any given model is reasoning about spacecraft behavior or confabulating plausible-sounding nonsense.

AstroMind is our answer to that gap. It converts high-fidelity astrodynamics simulations, constrained by real observational data, into verifiable reasoning problems spanning intent inference, maneuver parameter estimation, and threat assessment \cite{RAND_AI_SDA_2024}. Noise, sensor imperfections, and multi-source intelligence of varying credibility are woven into each scenario to reflect the actual information environment analysts face.

In this paper, we use AstroMind to make spacecraft behavior reasoning a reproducible LLM evaluation problem. The benchmark links physics-based simulation, observation-constrained scenarios, and multi-source textual evidence to tasks that require both numerical consistency and strategic interpretation. Our experiments then compare open-weight models across task accuracy, physical parameter estimation, and judged reasoning quality, showing that these capabilities do not always move together.

The core contributions of this work are:

\begin{itemize}
    \item A domain-specific benchmark for astronautics that asks whether LLMs can infer \textit{strategic intent} from multi-source observational evidence, rather than merely detect that an event occurred.

    \item A four-layer generation framework that ties professional astrodynamics simulation to real-world observational data, so each benchmark instance is answerable to both the physics and the operational record.

    \item A multi-dimensional task system spanning intent inference, parameter prediction, and threat assessment --- covering the full analytical chain from observation to judgment.

    \item \textbf{A hybrid semantic-numeric evaluation protocol (HSNE)} that separates task typing, semantic extraction, and numeric aggregation, making scoring robust to the linguistic variability of free-form LLM outputs.

    \item \textbf{An ablation of structured reasoning scaffolding} via the Reasoning Loop, showing that its effect is capability-sensitive: it produces improvements across all tested 8B models, with larger gains for models that can already track physical constraints.
\end{itemize}

The result is a framework that forces the question into the open: is the model reasoning about spacecraft behavior, or generating physically incoherent text that sounds right?

\section{Related Work}
\subsection{Computational Methods in Space Domain Awareness}

\subsubsection{Signal Processing and Its Limits}

The standard toolbox for spacecraft maneuver detection is built on residual analysis: compare a spacecraft's predicted position against observed measurements, and flag a maneuver when the gap exceeds a threshold \cite{Goff_2015_Real}. Kalman filters are the workhorse here \cite{Tapley_Orbit_2004}, and they work well for what they are designed to do. The trouble is distinguishing a real maneuver from sensor noise --- a distinction that drives persistent false-alarm problems in operational systems.

More recent work has pushed residual analysis further, applying it to impulsive maneuver estimation from optical \cite{Pastor_2022_Optical} and radar \cite{Porcelli_2022_Radar} survey data, and extending it to stochastic hybrid system formulations with Sequential Monte Carlo filtering \cite{Escribano_2022_Hybrid}. But they share the same fundamental character: they are reactive instruments that record \textit{what} a spacecraft did, not tools that can say \textit{why} it did it. Supervised ML models --- SVMs, random forests, deep neural networks --- improved accuracy and cut false positives \cite{fe2025}, yet inherited the same limitation. They classify statistical patterns in time-series data; they do not interpret intentions.

\subsubsection{AI for Autonomous Spacecraft Operations}

A parallel thread has applied AI to spacecraft guidance, navigation, and control (GNC), especially for autonomous rendezvous and active debris removal. Work in visual perception, including SpaceSeg \cite{liu2025spaceseg}, has shown that vision foundation models can achieve high-precision segmentation of on-orbit targets, enabling detailed state extraction from complex orbital scenes. Reinforcement learning has then been used to build control policies that map such perceptual inputs directly to actuator commands \cite{Oliveira2025OrbitZooRO}, establishing that fully autonomous space agents are technically feasible.

The critical distinction is that these systems are designed for ego-centric control: they optimize the behavior of a \textit{known} agent in a \textit{cooperative or well-defined} scenario. AstroMind targets a different layer --- inferring the intent of \textit{other} agents, potentially non-cooperative ones. Reasoning about what another spacecraft is doing, and why, is a prerequisite for robust autonomy in any multi-agent orbital environment. Existing GNC frameworks do not address it.

\subsection{LLM Benchmarks and the Aerospace Gap}
\subsubsection{From General to Specialized Evaluation}

Early LLM benchmarks like SuperGLUE \cite{Wang2019SuperGLUEAS} and MMLU \cite{Hendrycks_MMLU_2021} established general-purpose evaluation paradigms and revealed that broad language competence was advancing rapidly. As models matured, the field shifted toward domain-specific tests: MultiMedQA \cite{singhal2023large} for clinical reasoning, LegalBench \cite{Guha_LegalBench_2023} for statutory interpretation, HumanEval \cite{chen2021evaluating} for code generation. The pattern was consistent --- general benchmarks could not capture the failure modes that matter in expert domains, so purpose-built alternatives emerged.

\subsubsection{The Missing Case: Astrodynamics}

Aerospace engineering has been left out of this trend. As Table~\ref{sample-table} shows, no existing benchmark targets the physics-constrained setting of orbital mechanics and spacecraft operations. Without a standardized evaluation tool, any claim about LLM capability in this domain --- whether from a vendor or a research lab --- rests on anecdote rather than measurement. AstroMind was built to close that gap.

\begin{table*}[t]
\caption{Comparison of Mainstream LLM Reasoning Benchmarks}
\label{sample-table}
\centering
\small
\begin{tabularx}{\textwidth}{
    >{\raggedright\arraybackslash}p{3.2cm}  % 第一列
    >{\raggedright\arraybackslash}p{2.5cm}  % 第二列
    >{\raggedright\arraybackslash}p{3.5cm}  % 第三列
    >{\raggedright\arraybackslash}X         % 第四列
}
\toprule
% --- 表头区域：只在这里使用 Small Caps (\textsc) ---
\textsc{Benchmark Name} & \textsc{Primary Domain} & \textsc{Core Task} & \textsc{Reasoning Focus}\\
\midrule
% --- 内容区域：使用正常字体 ---
MMLU \cite{Hendrycks_MMLU_2021} & General Knowledge & Multiple-Choice Q\&A & Multitask Knowledge Application \\
\addlinespace
SuperGLUE \cite{Wang2019SuperGLUEAS} & Natural Language & Language Understanding & Inference, Contextual Understanding \\
\addlinespace
HumanEval \cite{chen2021evaluating} & Software Eng. & Code Generation & Algorithmic Logic \\
\addlinespace
MultiMedQA \cite{singhal2023large} & Medicine & Medical Q\&A & Clinical Knowledge \& Diagnosis \\
\addlinespace
LegalBench \cite{Guha_LegalBench_2023} & Law & Legal Task Classification & Legal Reasoning \& Interpretation \\
\addlinespace
% 你的方法：加粗即可，不需要 Small Caps
\textbf{AstroMind (This Work)} & \textbf{Aerospace Eng.} & \textbf{Intent Inference, Decision Analysis} & \textbf{Causal Reasoning Under Physical Constraints} \\
\bottomrule
\end{tabularx}
\end{table*}

\subsection{Reasoning Paradigms Relevant to AstroMind}

\subsubsection{Structured Reasoning and Chain-of-Thought}

Chain-of-Thought (CoT) prompting \cite{wei2022chain} established that asking a model to show its work substantially improves performance on multi-step problems. Tree-of-Thoughts (ToT) \cite{yao2023tree} extended this further, allowing parallel exploration of reasoning branches. AstroMind's ``Reasoning Loop'' mechanism and reasoning-chain evaluation metrics are built on this intuition --- the goal is not just to score final answers, but to assess whether the intermediate reasoning reflects coherent physical understanding.

\subsubsection{Physics-Informed and Multimodal Reasoning}

Standard LLMs tend to ignore physical constraints when they conflict with statistical patterns \cite{lewkowycz2022solving}, making them unreliable on problems with hard physical laws \cite{karniadakis2021physics}. AstroMind scenarios compound this challenge by combining numerical orbital time-series with natural-language intelligence reports, requiring a model to handle both data modalities and keep them mutually consistent. This makes the benchmark a concrete testbed for physics-informed, multimodal reasoning under real constraints.

The field has moved from general-purpose ``g-factor'' scoring \cite{spearman1904general} toward domain-specific tests of distinct capabilities \cite{gardner1983frames}. AstroMind brings that approach to the space domain, where the capability that matters is integrating physical dynamics with strategic inference---something no prior benchmark has measured.

\section{Method}

AstroMind uses a four-tier pipeline to convert astrodynamic simulations and observational data into structured evaluation tasks for large language models (Fig.~\ref{fig:framework}). Each tier addresses a distinct transformation: from physics to measurements, from measurements to noisy multi-source data, from data to prompt-formatted tasks, and from tasks to scored results.

\subsection{Tier 1: Dynamics Simulation and Observational Data}

\subsubsection{\textbf{Physics-based simulation}}
We built the simulation layer on Poliastro, an open-source orbital mechanics library (v0.17; note that the project was archived in 2023, but the propagation functionality used here remains fully functional). Trajectories are propagated with Cowell's method using Runge-Kutta integration, incorporating Earth's $J_2$ oblateness, an exponential atmospheric drag model, and impulsive maneuver events.

\subsubsection{\textbf{Real-world data integration}}
We supplemented simulations with ground-based azimuth and elevation angle time-series measurements to anchor synthetic scenarios in physical reality. Simulated trajectories generate predicted observation sequences, which are then aligned with the actual ground-truth data and iteratively adjusted until the residuals fall within acceptable bounds. This closed-loop fitting ensures that each scenario is physically self-consistent rather than purely synthetic.

\subsubsection{\textbf{Output format}}
Each scenario produces a time series of timestamps, state vectors (position $r(t)$ and velocity $v(t)$ in the J2000 inertial frame), and classical orbital elements ($a$, $e$, $i$, $\Omega$, $\omega$, $\nu$).

\subsection{Tier 2: Synthetic Data Generation}

This tier layers realistic sensor noise, missing data, and textual intelligence reports on top of the clean simulations, so that LLMs must reason under the kinds of imperfect information present in real space surveillance.

\subsubsection{\textbf{Event taxonomy construction}} 

\begin{figure*}[t]
    \centering
    \includegraphics[width=0.98\textwidth]{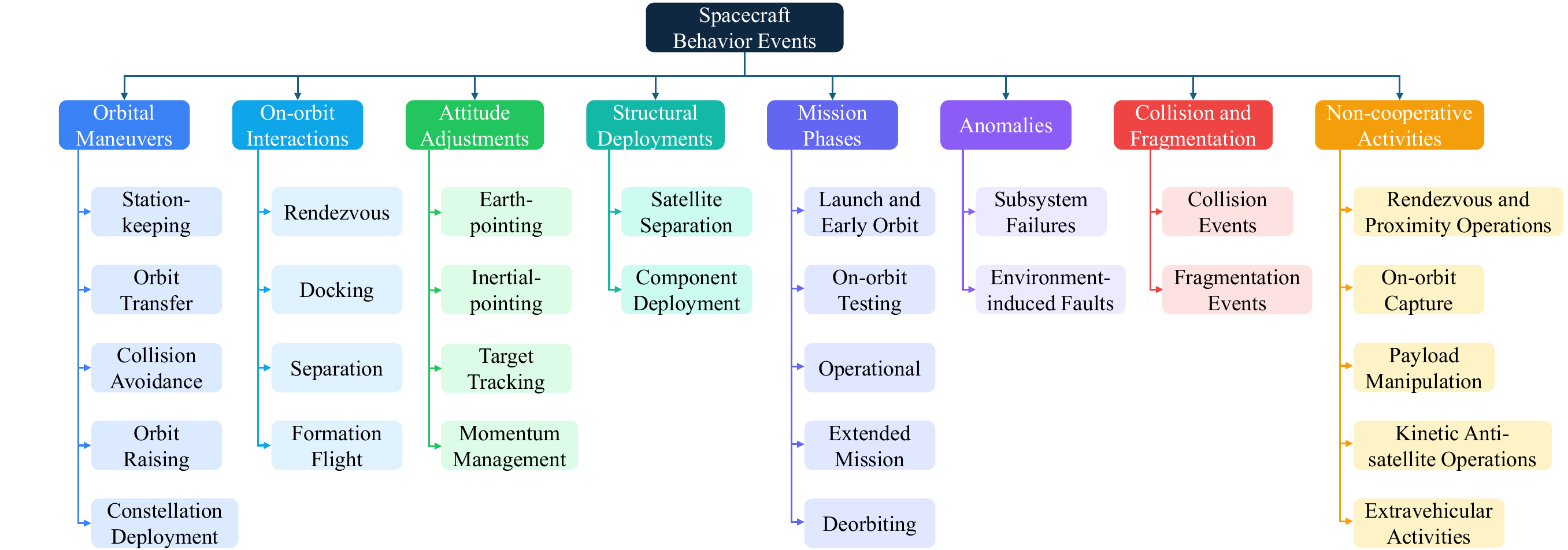}
    \caption{\textbf{Hierarchical Taxonomy of Spacecraft Behaviors in AstroMind.} Eight primary categories and 29 subcategories, derived from historical incidents and operational doctrine. The range runs from routine behaviors (Station-keeping, Deployment) to high-stakes non-cooperative activities (RPO, Kinetic ASAT)---each subcategory maps to a distinct, observable motion signature, so intent labels are physically anchored rather than arbitrary.}
    \label{fig:taxonomy}
\end{figure*}

We built the event taxonomy through a combined top-down and bottom-up process.

\textit{\textbf{Top-down}} (doctrinal alignment): Primary categories were aligned with Joint Publication 3-14: Space Operations \cite{JP3-14} and the CCSDS Orbit Data Messages standard \cite{CCSDS_ODM}. This grounds the taxonomy in recognized military doctrine and ground-station protocols, covering the full range from routine cooperative maintenance to potential counterspace threats.

\textit{\textbf{Bottom-up}} (data-driven refinement): We reviewed over 74 historical incidents---including the Cosmos 2251 collision and the Luch/Olymp-K RPO maneuvers---to identify the physical signatures that distinguish different behavior types. For Rendezvous and Proximity Operations (RPO) in particular, we cross-referenced behavioral criteria from the CONFERS standards \cite{CONFERS_2022} and SWF Global Counterspace Reports \cite{SWF_Counterspace2025}. This produced 29 subcategories (Figure~\ref{fig:taxonomy}) that map to distinct, observable motion patterns.

The result is a taxonomy that lets AstroMind test not just maneuver detection but intent inference, such as distinguishing ``Station-Keeping'' from ``Phasing for Interception'' based on subtle differences in burn vector and timing.

\subsubsection{\textbf{Benchmark dataset composition}}

\begin{figure*}[t]
    \centering
    \includegraphics[width=0.98\textwidth]{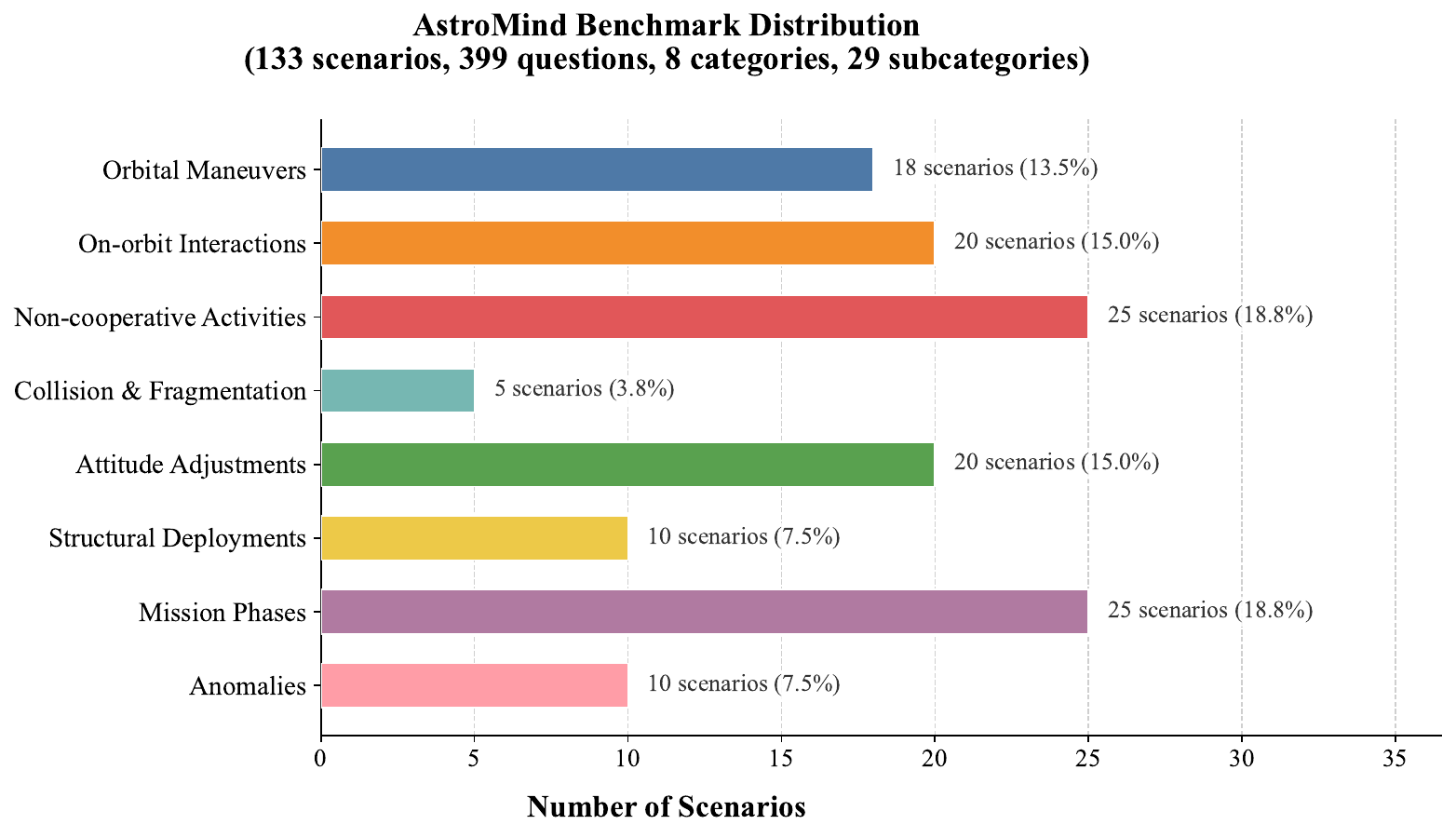}
    \caption{\textbf{AstroMind Benchmark Scenario Distribution.} Horizontal bar chart showing scenario counts across the 8 primary categories (133 scenarios, 399 questions total). Mission Phases and Non-cooperative Activities are jointly the largest categories (25 scenarios each, 18.8\%)---these are the behaviors where misclassification carries the highest operational cost, so their weight in the benchmark is deliberate. Each bar is colour-coded by category; percentages are annotated at the right.}
    \label{fig:dist}
\end{figure*}

The benchmark contains \textbf{133 scenarios and 399 benchmark questions} distributed across the 8 categories and 29 subcategories (Table~\ref{tab:dataset}, Figure~\ref{fig:dist}). In the released evaluation split analyzed here, these 399 questions break down into 66 intent-inference items, 241 parameter-estimation items, and 92 threat-assessment items. The benchmark is therefore not balanced one-question-per-dimension within the evaluated split, and the three task families should be interpreted against their own denominators rather than as if they were sampled in equal counts from every scenario.

Of the 66 intent-inference questions, 31 are \texttt{single\_choice\_with\_reasoning} (four options) and 35 are open-ended; parameter estimation and threat assessment questions are open-ended throughout. Both formats require explicit reasoning chains, not just a final answer.

The category distribution reflects SDA operational priorities. \textit{Mission Phases} and \textit{Non-cooperative Activities} are the largest categories (25 scenarios, 75 questions each; 18.8\% apiece), together accounting for 37.6\% of the benchmark. Mission Phases spans the spacecraft lifecycle from LEOP through deorbiting; Non-cooperative Activities covers the highest-stakes behaviors---Kinetic ASAT operations, RPO surveillance, on-orbit capture---where misclassification carries real strategic cost. \textit{On-orbit Interactions} and \textit{Attitude Adjustments} each contribute 20 scenarios (60 questions, 15.0\%). Smaller but operationally distinct categories---\textit{Collision \& Fragmentation} (5 scenarios, 3.8\%) and \textit{Anomalies} (10 scenarios, 7.5\%)---cover rare but consequential events. Most subcategories contain 5 scenarios (15 questions); two within Orbital Maneuvers have fewer than 5 scenarios---Orbit Transfer (1 scenario) and Station-keeping (2 scenarios)---reflecting the scarcity of unambiguous real-world exemplars.

% AstroMind Benchmark Composition Table
% 8 categories · 29 subcategories · 133 scenarios · 399 questions
\begin{table*}[t]
\caption{AstroMind Benchmark Composition: 8 Categories, 29 Subcategories, 133 Scenarios, and 399 Questions. Percentages are rounded and may not sum to exactly 100\%.}
\label{tab:dataset}
\centering
\small
\renewcommand{\arraystretch}{1.25}
\setlength{\tabcolsep}{5pt}
\begin{tabular}{@{}p{3.6cm} p{9.0cm} r r r@{}}
\toprule
\textbf{Category} & \textbf{Subcategories} & \textbf{Scenarios} & \textbf{Questions} & \textbf{Share} \\
\midrule
\textbf{Orbital Maneuvers}
  & Collision Avoidance,\ \ Orbit Raising,\ \ Constellation Deployment,\ \ Orbit Transfer,\ \ Station-keeping
  & 18 & 54 & 13.5\% \\
\textbf{On-orbit Interactions}
  & Rendezvous,\ \ Docking,\ \ Formation Flight,\ \ Separation
  & 20 & 60 & 15.0\% \\
\textbf{Non-cooperative Activities}
  & Rendezvous and Proximity Operations,\ \ On-orbit Capture,\ \ Payload Manipulation,\ \ Kinetic Anti-satellite Operations,\ \ Extravehicular Activities
  & 25 & 75 & 18.8\% \\
\textbf{Collision \& Fragmentation}
  & Collision Events,\ \ Fragmentation Events
  & 5  & 15 &  3.8\% \\
\textbf{Attitude Adjustments}
  & Earth-pointing,\ \ Inertial-pointing,\ \ Target Tracking,\ \ Momentum Management
  & 20 & 60 & 15.0\% \\
\textbf{Structural Deployments}
  & Component Deployment,\ \ Satellite Separation
  & 10 & 30 &  7.5\% \\
\textbf{Mission Phases}
  & Launch \& Early Orbit (LEOP),\ \ On-orbit Testing,\ \ Operational,\ \ Extended Mission,\ \ Deorbiting
  & 25 & 75 & 18.8\% \\
\textbf{Anomalies}
  & Subsystem Failures,\ \ Environment-induced Faults
  & 10 & 30 &  7.5\% \\
\midrule
\textbf{Total} & \textit{29 subcategories across 8 categories}
  & \textbf{133} & \textbf{399} & \textbf{100\%} \\
\bottomrule
\end{tabular}
\end{table*}

\subsubsection{\textbf{Observation noise modeling}}
We injected noise from three sources to simulate realistic sensor limitations: (1) zero-mean Gaussian white noise scaled by sensor type---optical or radar---with standard deviations of 1--10 arcseconds (angular) and 1--10 meters (range); (2) zenith-angle-dependent systematic biases approximating atmospheric refraction, most pronounced at low elevation; and (3) temporal jitter from signal propagation and processing delays. Observation sampling follows non-uniform intervals determined by satellite visibility windows, weather conditions, and station availability. Coverage gaps reflect the sparse geographic distribution of actual ground networks. In high-density scenarios such as formation flight or debris clouds, a stochastic dropout mechanism simulates sensor saturation.

\subsubsection{\textbf{Multi-source text generation}}
Each scenario also includes simulated textual intelligence: mission announcements, orbital anomaly reports, technical assessments, and news articles. These texts follow three design rules: they contain enough information to support reasoning but do not give away the answer; they span a credibility spectrum from official statements to social media rumors; and some are deliberately ambiguous or contradictory, requiring the model to reconcile conflicting evidence rather than simply extract a consensus claim.

\subsection{Tier 3: Scenario Abstraction and Task Construction}

This tier converts the numerical and textual data into prompt-formatted tasks that LLMs can reason over.

\subsubsection{\textbf{Prompt structure}}
Each prompt follows a four-part format: (1) \textit{background}---spacecraft identity, mission framework, and spatiotemporal context; (2) \textit{observational data}---serialized orbital elements or state vectors; (3) \textit{auxiliary information}---the multi-source text from Tier 2; (4) \textit{question}---the specific inference task with output format requirements. We use structured JSON with clear section headers, which improves both human readability and model format adherence.

\subsubsection{\textbf{Three reasoning dimensions}}

\textit{\textbf{Intent inference}} asks models to identify the behavioral purpose behind an observed maneuver sequence. A representative question: ``Given the observation sequence and intelligence, which action did Target A most likely execute? (A) station-keeping (B) phasing maneuver (C) rendezvous approach (D) anti-satellite attack. Select the most probable option and explain your reasoning, including key evidence and rationale for excluding alternatives.'' This tests evidence integration, causal reasoning, and space domain knowledge.

\textit{\textbf{Parameter estimation}} asks models to quantify the physical parameters of a maneuver. A representative question: ``Based on pre- and post-maneuver orbital elements, estimate the magnitude (m/s) and direction (unit vector in inertial frame) of the $\Delta v$ applied. Explain your methodology and assumptions.'' This tests numerical reasoning and orbital mechanics.

\textit{\textbf{Threat assessment}} asks models to evaluate the strategic implications of an observed behavior. A representative question: ``Assuming Target A belongs to Nation X and Target B to Nation Y, what does Target A's inferred intent imply for Target B's space asset security? Rate the threat level (low/medium/high) and analyze from both technical and strategic perspectives.'' This goes beyond physics to test contextual judgment and risk reasoning.

\subsubsection{\textbf{Difficulty levels}}
All scenarios in the current release involve multi-step reasoning chains, realistic sensor noise, and simulated intelligence reports that may be ambiguous or contradictory. Some problems are composite, combining all three dimensions sequentially (infer intent, estimate parameters, assess threat) to simulate a complete analytical workflow.

\begin{figure*}[t]
    \centering
    \includegraphics[width=0.9\textwidth]{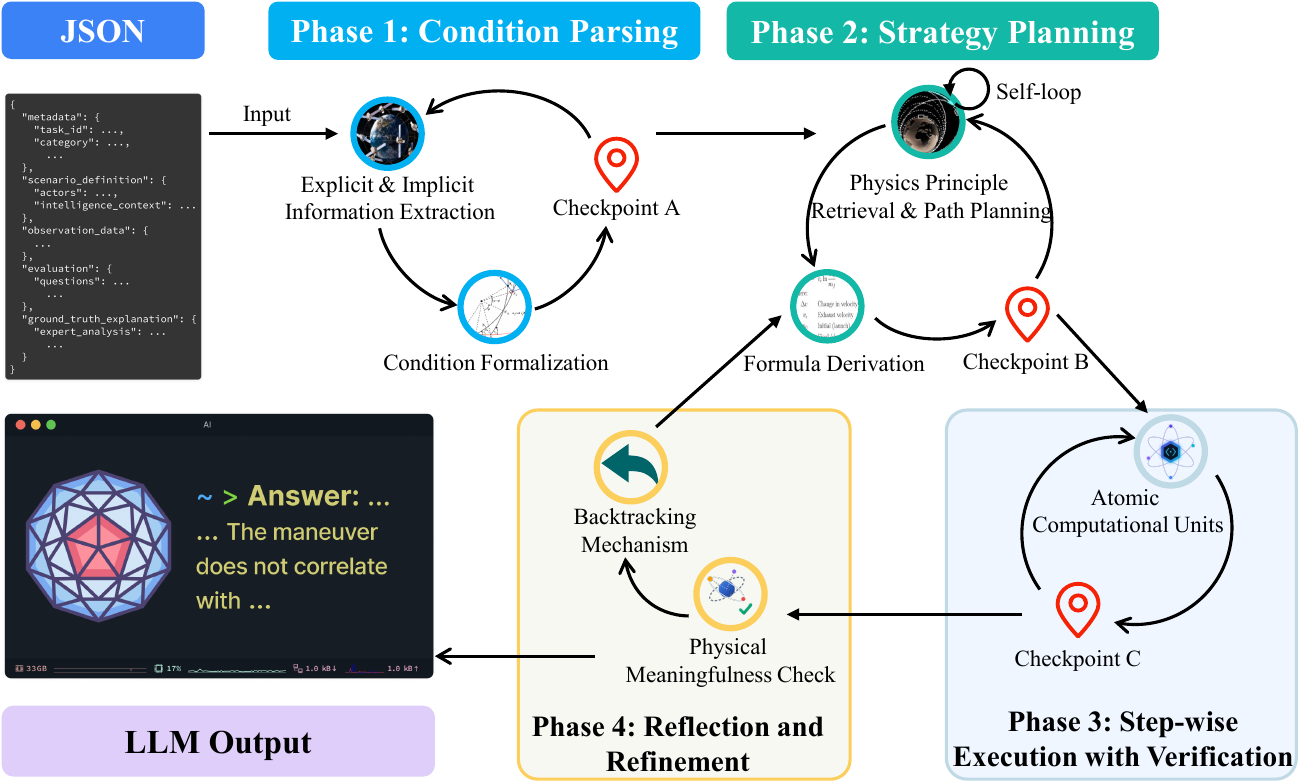}
    \caption{\textbf{The ``Reasoning Loop" Architecture.} An iterative inference scaffold that enforces physical consistency through four phases: (1) Condition Parsing---extracting explicit and implicit constraints; (2) Strategy Planning---deriving applicable formulas before computation; (3) Step-wise Execution with per-step unit/dimensionality checks; and (4) Reflection and Refinement, where a physical-plausibility check can trigger backtracking to earlier assumptions.}
    \label{fig:RL}
\end{figure*}

\subsection{Tier 4: Evaluation and Metrics}

\subsubsection{\textbf{Accuracy metrics}}
Task families are fixed before judging through dataset-exported question metadata and deterministic mapping rules, rather than inferred by the judge at evaluation time.

For multiple-choice and other classification tasks, we use standard accuracy:
$\text{Accuracy} = \frac{1}{N}\sum_{i=1}^{N}\mathbbm{1}[\hat{y}_i = y_i]$
where $N$ is total samples, $\hat{y}_i$ is the predicted label, $y_i$ is the ground-truth label, and $\mathbbm{1}[\cdot]$ is the indicator function.

For mixed-format intent questions, open-ended answers are first mapped by a judge model to the benchmark's canonical label set, then aggregated with multiple-choice items under the same accuracy formula. Because open-ended items pass through this additional judge-assisted mapping step, their effective error rate may differ from that of multiple-choice items; the aggregated accuracy figure should be treated as approximate rather than a strictly comparable value across formats.

For parameter estimation, we report relative error only on non-zero targets where both the predicted and reference scalars can be reliably parsed under the HSNE protocol (described below). We include the valid-sample count as ``Parsed N'' in the result tables because extraction coverage varies across models. In the released split, intent and threat accuracy are aggregated over 66 and 92 items respectively, whereas scalar parsing begins from 241 parameter-estimation items before model-specific extraction failures are applied. For $\Delta v$ magnitude:
\[
\text{RE}_{\Delta v} = \frac{|\hat{\Delta v} - \Delta v_{\text{true}}|}{|\Delta v_{\text{true}}|}.
\]
Directional and unit consistency are handled by the judge-assisted semantic checks and the physical-soundness rubric rather than reported as separate aggregate metrics.

\subsubsection{\textbf{Reasoning quality rubric}}
Beyond final-answer accuracy, we score the reasoning chain on three dimensions using a fixed DeepSeek-R1 judge model with deterministic JSON output:

\textit{\textbf{Logical coherence}} (5-point scale): whether causal links and argument structure are sound. 5 = fully coherent with clear derivations; 3 = followable but with obvious gaps; 1 = incoherent.

\textit{\textbf{Physical soundness}} (5-point scale): whether astrodynamic principles---Kepler's laws, conservation laws, orbital element interpretation---are applied correctly. 5 = fully correct; 3 = partially correct with clear misapplications; 1 = fundamental errors throughout.

\textit{\textbf{Reasoning completeness}} (5-point scale): whether all critical aspects are addressed. 5 = comprehensive; 3 = clear gaps in key considerations; 1 = severely incomplete.

\subsubsection{\textbf{Hybrid Semantic-Numeric Evaluation (HSNE)}}
Scoring free-form LLM outputs in a technical domain requires balancing two competing needs: flexibility to handle varied linguistic expression, and reliability to prevent stochastic errors from contaminating numeric aggregation \cite{Bhattarai2024LeveragingGF,gu2026surveyllmasajudge}. Rule-based extractors are brittle; purely generative evaluators introduce their own noise. HSNE decouples the two by separating semantic interpretation from arithmetic.

The protocol runs in three phases:
\begin{itemize}
    \item \textbf{Phase I: Task typing and semantic extraction.} Each sample's task family is fixed by benchmark metadata. The judge model then extracts structured representations from free-form answers (e.g., resolving ``approx.\ 12.5~km/s'' to $\langle\text{delta-v},\,12.5,\,\text{km/s}\rangle$) and compares qualitative conclusions against the reference answer.

    \item \textbf{Phase II: Deterministic computation.} A Python engine processes the structured outputs. Relative error and other quantitative metrics are computed by ordinary arithmetic, keeping numeric aggregation free from LLM stochasticity.

    \item \textbf{Phase III: Semantic alignment verification.} For qualitative tasks, the judge assesses whether the model's rationale matches the ground-truth annotation, producing per-sample binary correctness signals ($c_i \in \{0,1\}$) that feed into deterministic aggregation.
\end{itemize}
Separating task typing, semantic extraction, and numeric computation makes the evaluation both flexible enough to score unstructured outputs and stable enough to report reliable metrics.

\begin{figure*}[h]
    \centering
    \includegraphics[width=0.98\textwidth]{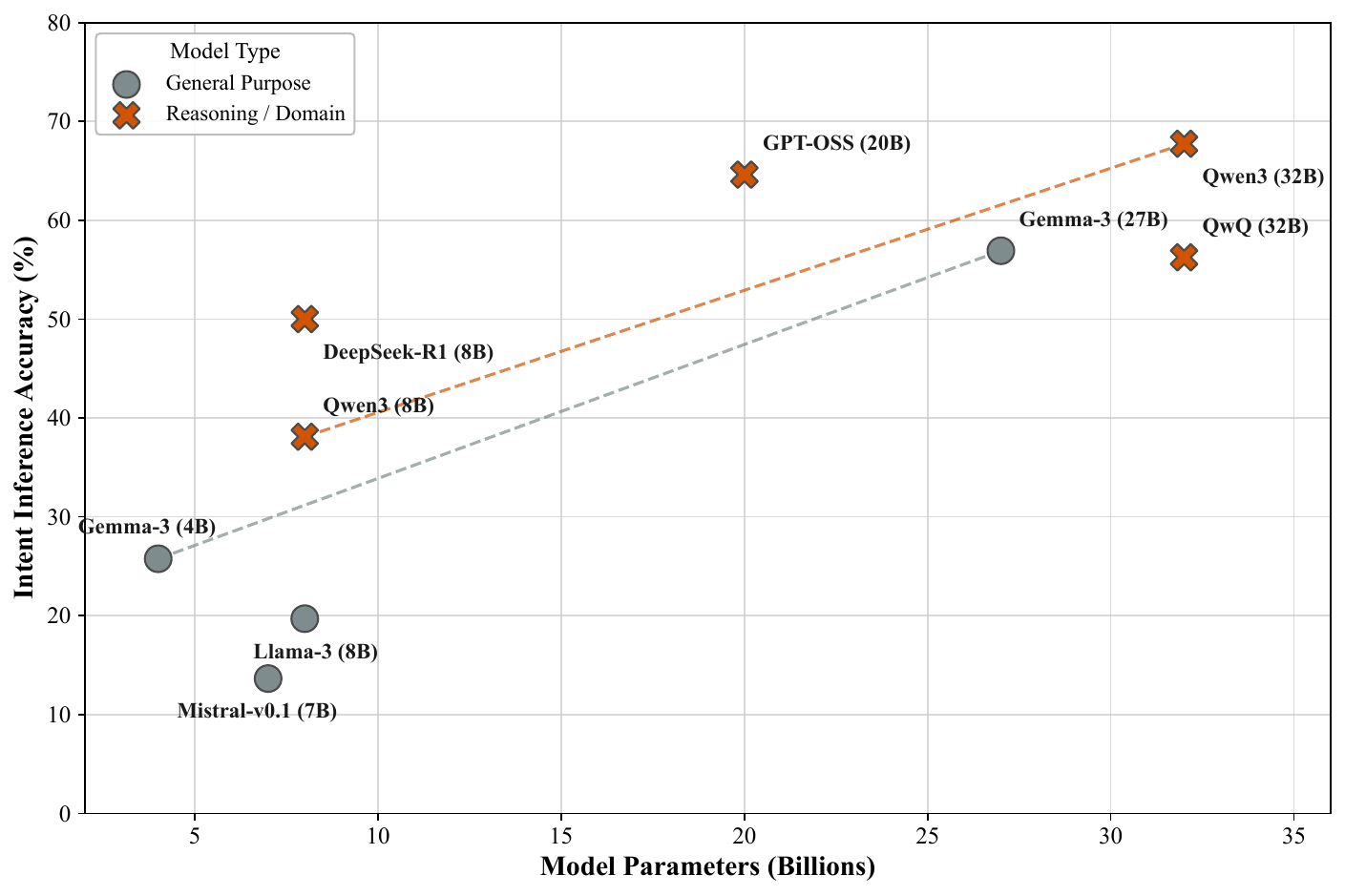}
    \caption{\textbf{Intent Inference Accuracy vs.\ Model Parameter Size.} Three patterns from Section~\ref{sec:experiments_analysis} are visible: training data composition matters (GPT-OSS at 20B is competitive with larger general-purpose models); a threshold-like scaling pattern is observed within the families tested here (Gemma and Qwen scaling trends); and reasoning-oriented models vary widely, with QwQ (32B) well ahead of smaller reasoning baselines.}
    \label{fig:RES}
\end{figure*}

\subsection{Investigating Structured Reasoning: The ``Reasoning Loop" Architecture}

Chain-of-Thought prompting works well in general settings, but whether it holds up under strict physics constraints is less clear. We introduce the Reasoning Loop to test this: an iterative inference framework grounded in metacognitive monitoring and P\'olya's problem-solving heuristics \cite{polya2014solve}, designed to make each reasoning step auditable. Unlike standard linear CoT, the loop folds back on itself, forcing the model to check its own work rather than just generate forward. We use it as a prompting scaffold in our ablation study, asking whether enforced verification actually helps current models or whether the added structure simply consumes context budget. The architecture (Figure~\ref{fig:RL}) has four phases:

\begin{itemize}
    \item \textbf{Phase 1: Condition Parsing.} The model extracts and formalizes all boundary conditions from the input---explicit physical quantities such as orbital elements, and implicit constraints such as sensor field-of-view limits. Making these explicit up front reduces the chance that a critical parameter gets dropped mid-solution.

    \item \textbf{Phase 2: Strategy Planning.} Before doing any computation, the model must state its approach: which formulas apply (e.g., the vis-viva equation), why they apply, and what the computational path looks like. Writing this out first activates relevant domain knowledge and catches strategy-level errors before they propagate.

    \item \textbf{Phase 3: Step-wise Execution with Verification.} Each computational step follows a fixed three-part structure: (1) operation description, (2) numerical calculation, and (3) unit/dimensionality check. Verifying units at each step catches errors early rather than letting them compound through a long derivation.

    \item \textbf{Phase 4: Reflection and Refinement.} After reaching a preliminary conclusion, the model checks whether the result is physically plausible---for example, whether a $\Delta v$ of 10~km/s is reasonable for a chemical thruster. If it detects a contradiction or anomaly, a \textbf{backtracking mechanism} sends the model back to re-examine earlier assumptions, turning unidirectional generation into a genuine closed loop.
\end{itemize}

In principle, externalizing verification into explicit steps should reduce working-memory load by offloading intermediate results to the context window. For smaller models ($<$10B parameters), however, the added verbosity may do the opposite---filling the context budget with scaffold text and leaving less room for the actual reasoning. Section~\ref{sec:ablation_reasoning_loops} tests which effect dominates.

\section{Experiments and Analysis}
\label{sec:experiments_analysis}

Real SDA deployments operate under two hard constraints: data sovereignty policy bars cloud access, and satellite SWaP-C budgets cannot support server-class models \cite{Diana_OnOrbit_2024, Yin_OEC_2025}. The emerging \emph{mother--daughter} constellation pattern makes both constraints concrete---daughter satellites collect raw observations while the mother satellite runs inference locally, cutting ground-station latency without offloading sensitive data \cite{Guo_Fractionated_2009, Shi_SatelliteEdgeAI_2025}. Hardware sets the ceiling: space-grade AI accelerators today sustain useful throughput on 7B--13B parameter models; 20B is a credible near-term target; 32B is roughly where the hardware roadmap runs out \cite{Diana_OnOrbit_2024}. We therefore evaluate open-weight LLMs in the 4B--32B range---not as a claim that any 32B model is flight-ready, but as a practical upper bound on what compact on-orbit deployments might support.

We evaluate three model categories: general-purpose baselines (Llama-3, Mistral, Gemma); an instruction-tuned open-weight model (GPT-OSS \cite{openai_gpt_oss_2025}); and reasoning-enhanced models (DeepSeek-R1 and QwQ, with QwQ treated as a reasoning-specialized member of the Qwen family). A DeepSeek-R1 judge model is used only inside the offline scoring pipeline and is not a deployment candidate.

\subsection{Overall Performance: Training Composition and Task Specialization}

\begin{figure}[h]
    \centering
    \includegraphics[width=0.98\columnwidth]{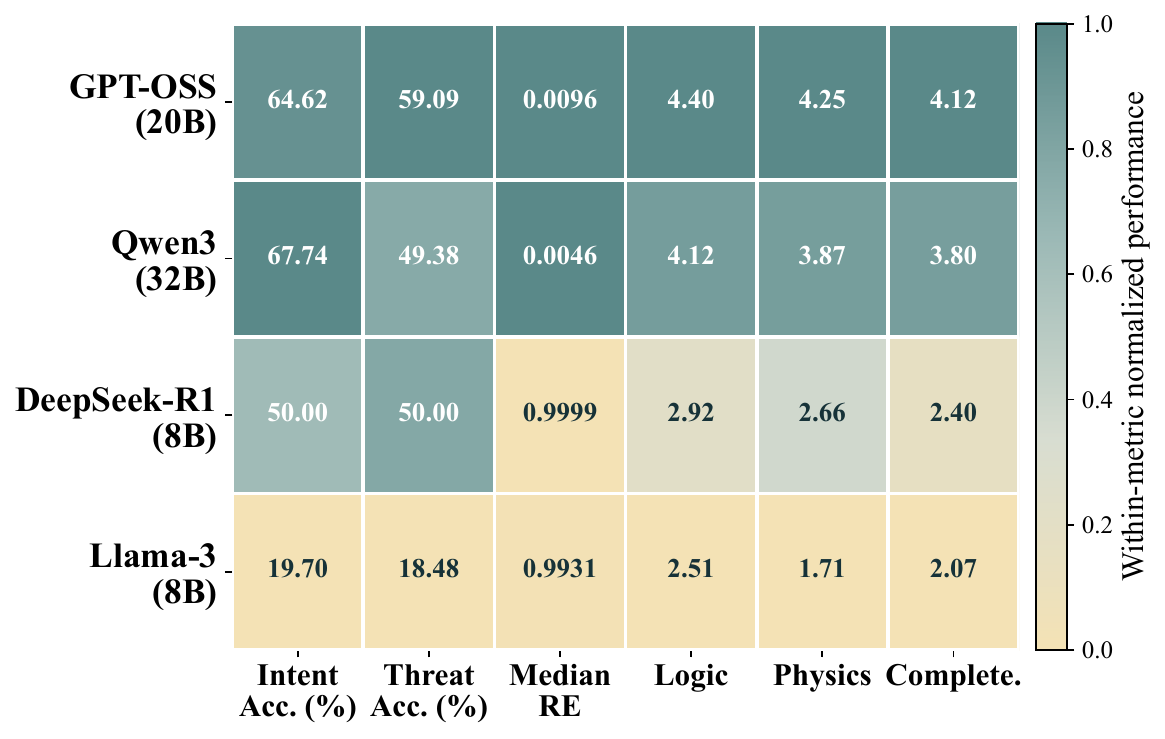}
    \caption{\textbf{Multi-dimensional Capability Profiling of Representative Models.} Heatmap of six metrics for four models: Intent Accuracy, Threat Accuracy, Median RE, Logic, Physics, and Completeness. Shading is column-normalized; raw values are annotated. Median RE should be read alongside ``Parsed N'' in Table~\ref{tab:main_results}, as extraction coverage differs by model. GPT-OSS (20B) leads the judged reasoning rubric; Qwen3 (32B) leads on intent accuracy and scores strongly on the rubric; DeepSeek-R1 (8B) metrics reflect only 99/399 items due to context-window saturation; Llama-3 (8B) is consistently weak across all metrics.}
    \label{fig:radar}
\end{figure}

Table~\ref{tab:main_results} shows results across all models. No single model wins on every metric; the benchmark separates models along three distinct axes: task-level accuracy (intent and threat), judged reasoning quality, and numerical parameter estimation. Qwen3 (32B) is the only model that performs strongly across all three---leading on intent accuracy (67.74\%), scoring highly on the reasoning rubric (Logic: 4.12), and achieving the lowest Median RE among models with substantial parsed coverage. QwQ (32B) leads on threat assessment (66.67\%) and posts the lowest absolute Median RE, but with narrower extraction coverage than GPT-OSS. GPT-OSS (20B) leads the reasoning rubric---logical coherence (4.40), physical soundness (4.25), reasoning completeness (4.12)---while remaining competitive on task metrics. Because the rubric is averaged over all tasks without re-weighting, and parameter-estimation items account for 241 of 399 evaluated items, the ``overall reasoning quality'' figure is dominated by performance on numerical estimation rather than reflecting equal contributions from all three task families.

Training data composition matters as much as scale. GPT-OSS (20B) matches or beats Gemma-3 (27B) despite having fewer parameters (Figure~\ref{fig:RES}), and Qwen3 (32B) substantially outperforms its 8B counterpart across every dimension. At the lower end, Mistral-v0.1 (7B) reaches only 13.64\% and Llama-3 (8B) 19.70\% on the mixed-format intent task; on the 31-item multiple-choice subset of that split, these levels fall at or below the random baseline. Median RE must be read alongside the ``Parsed N'' column: QwQ's parsed-subset Mean RE of 0.0779 reflects a highly skewed distribution---33 of 60 parsed samples (55\%) yield exactly zero relative error, while 8 samples with RE $\geq$ 0.1 pull the mean upward. This concentration of exact-zero errors likely reflects a parsing selectivity effect: HSNE successfully extracts a scalar from QwQ responses when the model states a clear numerical value, and QwQ frequently restates or closely echoes reference-range values provided in the scenario context. The Median RE of 0.0000 should therefore be interpreted as a lower bound on estimation accuracy for the parseable subset rather than a representative error rate across all parameter queries. Throughout this paper, Median RE is the primary reported metric; Mean RE appears here solely to characterize the distributional shape of QwQ's parsed subset. Scale helps, but only when the pre-training mixture includes sufficient STEM and domain-relevant content.

\subsection{Scaling Trends and Within-Family Capacity Effects}
Within the model families tested here, scale still matters---but the point at which gains become substantial appears to depend on training data composition. With only two scale points per family tested here, we cannot distinguish a true threshold from a steep linear trend; the pattern is consistent with a threshold but does not confirm one. In the Gemma family, intent accuracy scales with model size---Gemma-3 (4B) reaches only 25.76\% while Gemma-3 (27B) reaches 56.92\%---but Median RE does not follow: it remains high at 0.8511 even at 27B. The two tasks are pulling on different capabilities. Intent inference tracks model size because it draws on broad language and world knowledge; parameter estimation tracks training content because it needs reliable numerical reasoning---and Gemma's pre-training, at this scale, does not supply it. The Qwen3 family shows a cleaner pattern: scaling from 8B to 32B lifts intent accuracy from 38.10\% to 67.74\%, cuts Median RE from 0.0498 to 0.0046, and raises logical coherence from 3.39 to 4.12.

Models with richer STEM pre-training appear to reach comparable performance at smaller scale---Qwen3 (8B) already performs reasonably where Gemma-3 (27B) is still unreliable---suggesting that, in the families tested here, the point at which gains become substantial is not a fixed parameter count but varies with training data composition.

\begin{figure}[t]
    \centering
    \includegraphics[width=0.98\columnwidth]{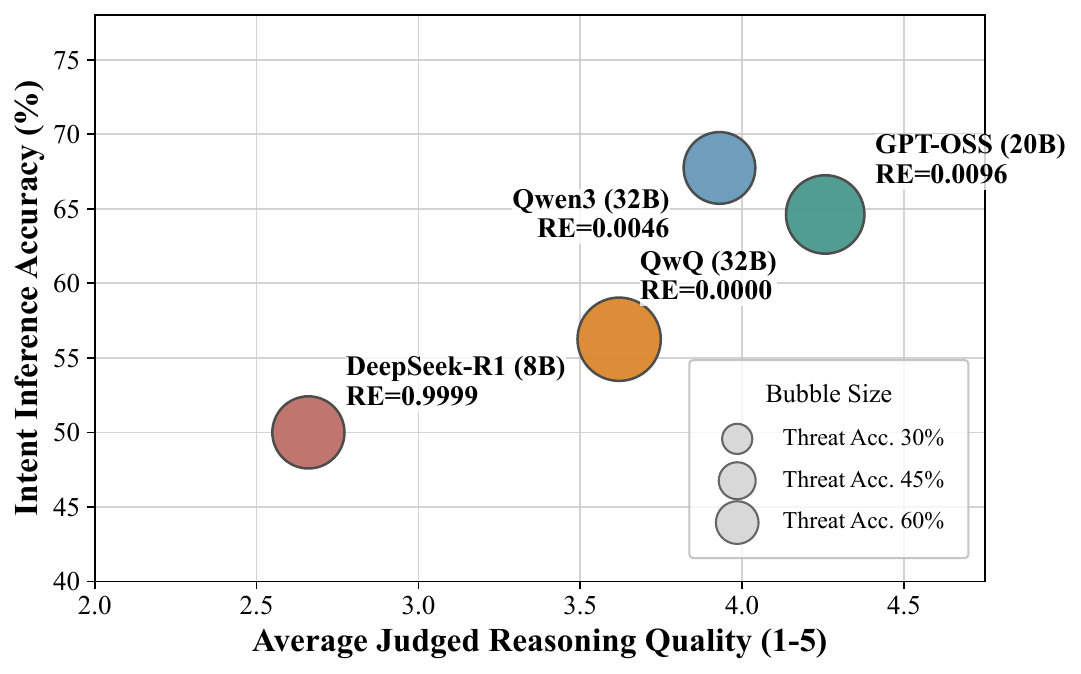}
    \caption{\textbf{Task Accuracy and Judged Reasoning Quality Among Reasoning-Oriented Models.} Horizontal axis: average judged score (logic, physics, completeness). Vertical axis: intent inference accuracy. Bubble size: threat assessment accuracy. Text labels: Median RE, computed on model-specific parsed subsets---read alongside ``Parsed N'' in Table~\ref{tab:main_results}. Judged reasoning quality and task accuracy are correlated but not the same: GPT-OSS (20B) leads the rubric; Qwen3 (32B) leads on intent accuracy; QwQ (32B) leads on threat accuracy and posts the lowest parsed-subset Median RE.}
    \label{fig:radar2}
\end{figure}

\begin{table*}[t]
\centering
\caption{\textbf{Performance on the AstroMind Benchmark.} Metrics: Intent Inference (Accuracy over scored items), Parameter Estimation (Median RE), Threat Assessment (Accuracy over scored items), and judge-model Reasoning Quality (5-point scale). Median RE is computed only on parameter-estimation items where both predicted and reference scalars are successfully extracted; ``Parsed N'' gives the corresponding count out of 241 parameter-estimation items. Items with empty model responses are excluded from scoring and counted separately; DeepSeek-R1 (8B) returned non-empty responses for only 99/399 items, so its metrics reflect a narrow subset. In all other columns, \textbf{bold} and \underline{underline} mark the best and second-best values.}
\label{tab:main_results}
\resizebox{\textwidth}{!}{%
\begin{tabular}{lcccccccc}
\toprule
\multirow{2}{*}{\textbf{Model}} & \multirow{2}{*}{\textbf{Size}} & \textbf{Intent Inference} & \multicolumn{2}{c}{\textbf{Parameter Est.}} & \textbf{Threat Assess.} & \multicolumn{3}{c}{\textbf{Reasoning Quality (5-point scale)}} \\
\cmidrule(lr){3-3} \cmidrule(lr){4-5} \cmidrule(lr){6-6} \cmidrule(lr){7-9}
 &  & Accuracy ($\uparrow$) & Median RE ($\downarrow$) & Parsed N & Accuracy ($\uparrow$) & Logic & Physics & Completeness \\
\midrule
\multicolumn{9}{l}{\textit{General Purpose Baselines}} \\
Gemma-3 & 4B & 25.76\% & 0.9999 & 115/241 & 26.09\% & 2.61 & 1.93 & 2.43 \\
Mistral-v0.1 & 7B & 13.64\% & 0.9984 & 99/241 & 19.57\% & 2.40 & 1.74 & 1.88 \\
Llama-3 & 8B & 19.70\% & 0.9931 & 125/241 & 18.48\% & 2.51 & 1.71 & 2.07 \\
Gemma-3 & 27B & 56.92\% & 0.8511 & 135/241 & 46.74\% & 3.73 & 3.21 & 3.39 \\
\midrule
\multicolumn{9}{l}{\textit{Reasoning \& Domain-Capable Models}} \\
DeepSeek-R1$^\dagger$ & 8B & 50.00\% & 0.9999 & 5/241 & 50.00\% & 2.92 & 2.66 & 2.40 \\
Qwen3 & 8B & 38.10\% & 0.0498 & 85/241 & 45.45\% & 3.39 & 3.04 & 2.89 \\
GPT-OSS & 20B & \underline{64.62\%} & 0.0096 & 136/241 & \underline{59.09\%} & \textbf{4.40} & \textbf{4.25} & \textbf{4.12} \\
Qwen3 & 32B & \textbf{67.74\%} & 0.0046 & 111/241 & 49.38\% & \underline{4.12} & \underline{3.87} & \underline{3.80} \\
QwQ & 32B & 56.25\% & 0.0000 & 60/241 & \textbf{66.67\%} & 3.73 & 3.56 & 3.57 \\
\bottomrule
\end{tabular}%
}
\vspace{4pt}

\footnotesize $^\dagger$DeepSeek-R1 (8B) returned non-empty responses for only 99/399 items (75\% empty rate); metrics reflect only the attempted subset and are not directly comparable to models with near-full coverage.
\vspace{-6pt}
\end{table*}

\subsection{Subtask Imbalance and Reasoning-Outcome Decoupling in Reasoning-Focused Models}
\label{tax}
The reasoning-focused models reveal two patterns that do not fit a simple story about chain-of-thought helping or hurting.

\textbf{1. Correct answers and high-rated explanations come apart.} QwQ (32B) leads on threat accuracy (66.67\%) and posts the lowest parsed-subset Median RE, yet its judged reasoning scores---Logic: 3.73, Physics: 3.56, Completeness: 3.57---fall below GPT-OSS (20B). Qwen3 (32B) leads on intent accuracy (67.74\%) while also scoring strongly on the rubric (Logic: 4.12). A model can reach the right answer without producing the clearest written derivation, and vice versa; near-zero numerical error and high rubric scores measure different aspects of model capability and do not always move together.

\textbf{2. Coverage gaps from context-window saturation limit interpretation.} DeepSeek-R1 (8B) returned non-empty responses for only 99 of 399 items under the baseline condition (4,096-token context window). As a reasoning-oriented model, its extended internal thinking traces consumed the context budget before a final answer could be generated, resulting in truncated outputs. Doubling the context window to 8,192 tokens under the Loop condition increased valid response coverage to 248/399 (62\%), consistent with this interpretation. Reported scores for the baseline condition---50.00\% intent accuracy, 50.00\% threat accuracy, Logic: 2.92---reflect only the narrow subset the model successfully completed and are not directly comparable to models with near-full coverage.

\begin{figure*}[t]
    \centering
    \includegraphics[width=0.98\textwidth]{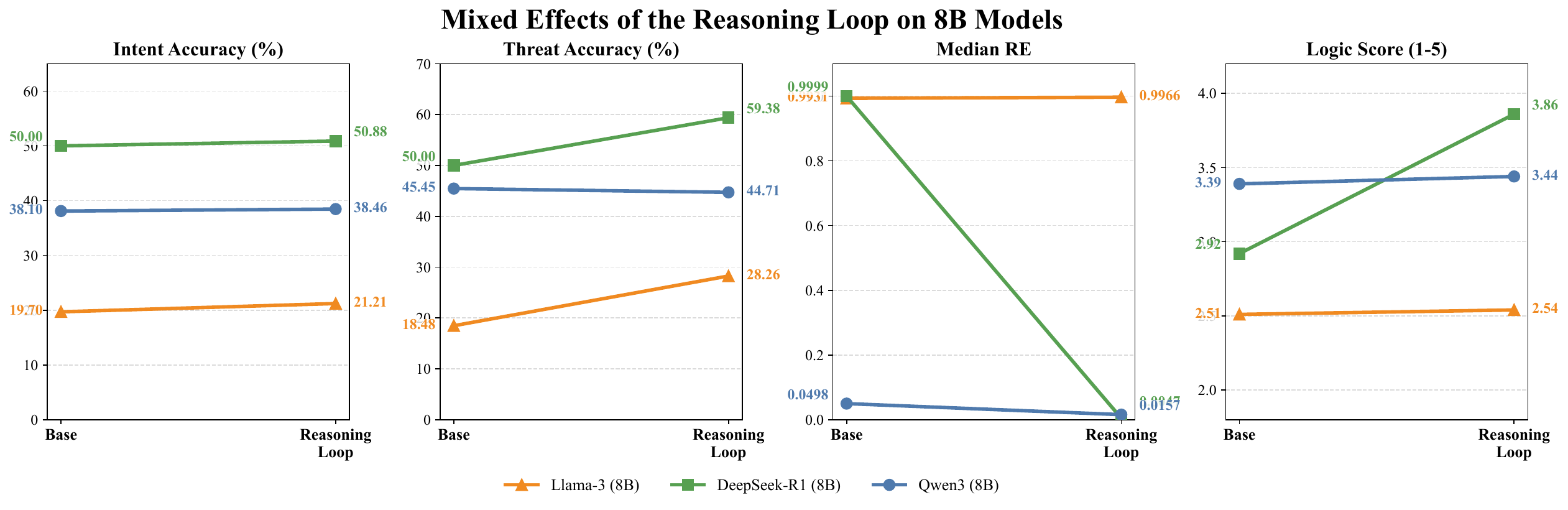}
    \caption{\textbf{Mixed Effects of the Reasoning Loop on 8B Models.}
Raw-value trajectories across four metrics (Intent Accuracy, Threat Accuracy, Median RE, Logic Score) with and without the Reasoning Loop. Qwen3~($\bullet$), DeepSeek-R1~($\blacksquare$), and Llama-3~($\blacktriangle$) show mixed effects: most metrics improve, but Llama-3's Median RE rises slightly (0.9931$\to$0.9966) and Qwen3's Threat Accuracy dips marginally (45.45\%$\to$44.71\%). Gains for DeepSeek-R1 partly reflect increased response coverage under the larger context window. Qwen3 and DeepSeek-R1 show stronger gains on reasoning quality than Llama-3.}
    \label{fig:AS}
\end{figure*}

\section{Ablation Study: Mixed Effects of Reasoning Loops}
\label{sec:ablation_reasoning_loops}

\subsection{Hypothesis: Structured Scaffolding Is Capability-Sensitive}
\label{sec:illusion_hypothesis}
Can prompting alone close the capability gap in compact models ($<$10B parameters)? We test this by applying the Reasoning Loop to three 8B models and measuring whether enforced step-by-step derivation and self-verification actually improves outcomes. Our expectation going in was that the loop might polish surface signals---making reasoning traces look more structured---while adding sequence-length and constraint-tracking demands that smaller models struggle to sustain, ultimately widening the gap between linguistic plausibility and numerical correctness.

\subsection{Results: Gains Are Model-Dependent}
\label{sec:alignment_execution_tradeoff}
The Reasoning Loop produces improvements across all three 8B models tested (Table~\ref{tab:reasoning_loop}, Figure~\ref{fig:AS}), though the magnitude and nature of gains differ substantially.

\begin{table}[h]
\centering
\caption{\textbf{Impact of the Reasoning Loop on 8B Models.} Baseline vs.\ loop-augmented performance on intent inference, threat assessment, parameter estimation, and judge-model logic score. Median RE follows the same scalar-extraction protocol as Table~\ref{tab:main_results}; Parsed N differs between settings and is listed in the footnote. For all other metrics, \textbf{bold} marks the better value within each model; ties are left unbolded. DeepSeek-R1 (8B) had substantially more non-empty responses under the Loop condition (248 vs.\ 99), so its improvements reflect both prompt effects and increased response coverage.}
\label{tab:reasoning_loop}
\resizebox{\columnwidth}{!}{%
\begin{tabular}{l|cc|cc|cc|cc}
\toprule
\multirow{2}{*}{\textbf{Model (8B)}} & \multicolumn{2}{c|}{\textbf{Intent Acc ($\uparrow$)}} & \multicolumn{2}{c|}{\textbf{Threat Acc ($\uparrow$)}} & \multicolumn{2}{c|}{\textbf{Median RE ($\downarrow$)}} & \multicolumn{2}{c}{\textbf{Logic Score ($\uparrow$)}} \\
 & Base & Loop & Base & Loop & Base & Loop & Base & Loop \\
\midrule
Llama-3     & 19.70\% & \textbf{21.21\%} & 18.48\% & \textbf{28.26\%} & 0.9931 & 0.9966 & 2.51 & \textbf{2.54} \\
DeepSeek-R1$^\dagger$ & 50.00\% & \textbf{50.88\%} & 50.00\% & \textbf{59.38\%} & 0.9999 & \textbf{0.0047} & 2.92 & \textbf{3.86} \\
Qwen3       & \textbf{38.10\%} & 38.46\% & \textbf{45.45\%} & 44.71\% & 0.0498 & \textbf{0.0157} & 3.39 & \textbf{3.44} \\
\bottomrule
\end{tabular}%
}
\vspace{2pt}

\footnotesize\emph{Parsed N (Base/Loop) for Median RE}: Llama-3 = 125/108, DeepSeek-R1 = 5/51, Qwen3 = 85/89.\\
\footnotesize $^\dagger$DeepSeek-R1 (8B): 99/399 non-empty responses (Base) vs.\ 248/399 (Loop); metrics are not directly comparable across conditions.
\end{table}

\subsubsection{Reasoning Quality Gains Are Selective}
\label{sec:qualitative_gain}
Logic scores rise for Qwen3 (3.39 $\to$ 3.44) and DeepSeek-R1 (2.92 $\to$ 3.86), reflecting more internally consistent reasoning traces. Llama-3 rises marginally from 2.51 to 2.54. The loop produces better-structured outputs only when the base model already has enough task competence to use the scaffold productively.

\subsubsection{Task-Level Effects Are Not Uniform}
\label{sec:quantitative_loss}
Task metrics follow the same split. Qwen3 shows mixed results: intent accuracy improves marginally (38.10\% $\to$ 38.46\%), Median RE improves substantially (0.0498 $\to$ 0.0157), while threat accuracy dips slightly (45.45\% $\to$ 44.71\%). DeepSeek-R1 improves substantially on the items it attempted (N=248 with Loop vs.\ N=99 without): intent accuracy 50.00\% $\to$ 50.88\%, threat accuracy 50.00\% $\to$ 59.38\%, Median RE 0.9999 $\to$ 0.0047; the higher coverage under the Loop condition means these improvements reflect both prompt effects and reduced generation failures. Llama-3 shows modest gains: intent accuracy 19.70\% $\to$ 21.21\%, threat accuracy 18.48\% $\to$ 28.26\%, Median RE 0.9931 $\to$ 0.9966. There is no uniform collapse---but there is also no universal gain.

\subsection{Discussion: Capability-Sensitive Scaffolding Rather Than Universal Collapse}
\label{sec:discussion_overload}
The Reasoning Loop is not a universal failure mode, nor a panacea. Across all three models tested, it produces improvements on most metrics---but the size and consistency of gains differ. For Qwen3 and DeepSeek-R1, enforced decomposition and self-checking improve both reasoning quality and task outcomes. For Llama-3, gains are more modest and uneven, suggesting that the scaffold's benefit scales with the base model's existing capacity to track physical constraints.

\paragraph{Implication.}
The practical capacity threshold for structured scaffolding is not a hard cutoff at 8B; in the models tested here, it varies within the same size class depending on training. Prompt engineering alone is unlikely to erase core capability gaps in physics-constrained tasks, though it can yield meaningful gains for some compact models. Robust performance will likely require either \emph{capacity-aware} prompting that controls verbosity and output format, or training-time interventions---fine-tuning, verifier-guided training, tool-augmented supervision---that internalize constraint tracking rather than delegating it to a longer prompt.

\section{Conclusion}
We asked whether large language models can go beyond detecting spacecraft maneuvers to inferring intent under realistic SDA conditions---with noisy observations, sparse coverage, and conflicting textual intelligence. AstroMind was built to test this: 133 scenarios, 399 questions, spanning intent inference, parameter estimation, and threat assessment across 29 behavior subcategories.

Three findings stand out. No single model dominates across all evaluation axes---task-level accuracy, judged reasoning quality, and numerical parameter estimation are partially independent, and Qwen3 (32B) is the only model to perform strongly on all three. Training data composition matters as much as scale: a well-targeted 20B model can outperform a generic 27B one on reasoning quality, and Gemma-3 (27B)'s high intent accuracy alongside persistently high Median RE shows that scale alone does not carry over to parameter estimation. Intent inference and parameter estimation draw on different capabilities---the former scales more readily with model size and general language knowledge, while the latter requires numerical reasoning that not all architectures develop at the same rate. Structured prompting gains scale with base capability: the Reasoning Loop improves all tested 8B models, with larger gains for those that already track physical constraints, though context-window effects partially confound gains for reasoning-oriented models.

The benchmark has real limits worth naming plainly. It is simulation-grounded and uses curated text, so it misses operational sensor noise, adversarial deception, and rare failure modes. The intent taxonomy covers a fixed set of maneuver classes, and generalization to longer-horizon campaigns has not been tested. Qualitative scoring relies on a single judge model, and Median RE is sensitive to parse coverage; at 66 intent-inference and 92 threat-assessment items, small accuracy differences between models are not statistically decisive and should be read as indicative rankings. The most glaring gap is the absence of a human-expert baseline. We do not know where these models sit relative to a trained analyst working the same problems, and that comparison must happen before the results can inform deployment decisions. On the research side, the field needs training-time methods that internalize physical constraint tracking---not more scenario types, and not prompt scaffolding. Scaffolding can mask a capability gap; it cannot close one.

\bibliographystyle{IEEEtran}
\bibliography{IEEEabrv,main}

\end{document}